\RequirePackage{fix-cm}
\documentclass[smallextended]{svjour3}       
\smartqed  
\usepackage{graphicx}
\usepackage{amsmath}
\usepackage{color}
\begin{document}

\title{Orientation Aware Weapons Detection In Visual Data : A Benchmark Dataset
}


\author{N.U.Haq \and M.M.Fraz \and T.S.Hashmi \and M.Shahzad 
}


\institute{Nazeef Ul Haq \at
              National University of Sciences and Technology (NUST), Islamabad, Pakistan \\
              \email{nhaq.mscs18seecs@seecs.edu.pk}           
           \and
           Muhammad Moazam Fraz \at
              The Alan Turing Institute, 96 Euston Rd, London NW1 2DB, United Kingdom
              \\
              mfraz@turing.ac.uk (M.M. Fraz)
}

\date{Received: date / Accepted: date}

\maketitle

\begin{abstract}
Automatic detection of weapons is significant for improving security and well being of individuals, nonetheless, it is a difficult task due to large variety of size, shape and appearance of weapons. View point variations and occlusion also are reasons which makes this task more difficult. Further, the current object detection algorithms process rectangular areas, however a slender and long rifle may really cover just a little portion of area and the rest  may contain unessential details. To overcome these problem, we propose a CNN architecture for Orientation Aware Weapons Detection, which provides oriented bounding box with improved weapons detection performance. The proposed model provides orientation not only using angle as classification problem by dividing angle into eight classes but also angle as regression problem. For training our model for weapon detection a new dataset comprising of total 6400 weapons images is gathered from the web and then manually annotated with position oriented bounding boxes. Our dataset provides not only oriented bounding box as ground truth but also horizontal bounding box. We also provide our dataset in multiple formats of modern object detectors for further research in this area. The proposed model is evaluated on this dataset, and the comparative analysis with off-the shelf object detectors yields  superior  performance of proposed model, measured  with  standard evaluation  strategies. The dataset and the model implementation are made publicly available at this link: https://bit.ly/2TyZICF.
\end{abstract}

\section{Introduction}
\label{intro}
Weapon savagery incidents are largely spread across the globe what's more, are being seen at expanding recurrence \cite{moian2018walking}, \cite{morris2018mass}, \cite{Texas} and \cite{planty2013firearm}.  
Consistently, around a quarter-million individuals die due to firearm savagery \cite{BBC}. Steps for gun control don't appear to be compelling despite such an enormous number of deplorable occasions. Firearm brutality consistently covers the globe, and it is making unfavorable consequences for mankind. The issue should be tended to logically for the advancement of general well-being, a person's health, and security. Solid voices have been raised recently for logical information-supported exploration to forestall weapon viciousness \cite{jaffe2018gun} and to finance research for such tasks \cite{NYT}. Over the world, security agencies, government, and private institutions have expanded the applications of surveillance systems to protect the lives of people, secure buildings, and monitor the commercial area. As in this globalized world, there are more people to people contacts but at the same time, gun violence due to mutual hatred is also common. So, an efficient firearm detection method rooted in a surveillance system to avert gun violence or make a prompt response is unavoidable\cite{velastin2006motion} and \cite{ainsworth2002buyer}. Such steps will not only ensure security but also boost the economy due to low medical costs because of timely response due to early detection of weapons through an automated mechanism.
    
Previously, visual systems have been used for the identification of objects in images or videos. These systems can also be used for firearm detection due to the particular characteristics of these systems.  Systems like based on CNN \cite{liu2016ssd},\cite{ren2015faster},\cite{redmon2016you} has used for such purposes however, the generic version was not suitable for firearm detection objects\cite{hu2018sinet,li2017scale}, \cite{li2017perceptual}, \cite{liu2019towards}and \cite{zhang2016faster}. The reason for this is hiding in the chemistry of firearm objects like unflattering viewing angles, clutter, occlusions, shape, and size of firearms which make the detection more tricky than other objects such as vehicles, human faces, and airplanes. The existing systems have one limitation that is the use of axis-aligned windows in object detection systems. Most experts analyze the features of window and determine the presence of objects however, due to the small size of guns, the physically elongated structure of guns make window inefficient in case of firearm detection. This happens owing to a very low foreground to background ratio where the weapon behaves like foreground and window with all its features other than weapon like a background. For example, when a person is carrying a gun and the window is busy detecting the gun, at the same time it will also detect the person who is carrying the gun. This amalgamated mixture makes the information difficult for the classifier to differentiate between a required object and background information. Zhou et al. proposed a feature detection method recently which is orientation aware, particularly for handling planer rotations however it could not prove efficient for handling clutter\cite{zhou2017oriented}. Some methods used a mechanism where angles were made a part of an anchor which resulted in an increased number of anchors that requires to be categorized for every object and proposal\cite{azimi2018towards},\cite{ma2018arbitrary}. The drawback lies in a large number of anchors because it is computationally ineffective. Moreover, these methods need to train region proposals on oriented bounding boxes but presently almost all data sets have horizontal bounding boxes as ground truth. There are very few data sets that have an oriented bounding box as ground truth. Our proposed method in this paper required oriented bounding box for training. 
     
In this research work, we propose an orientation-aware weapon detection algorithm in visual data which not only improves the detection of objects but also gives information about the orientation of objects. We also prepared a new dataset comprises of 6400 weapon images. Our dataset has an oriented bounding box as ground truth with angle information. According to our best knowledge, no comprehensive dataset related to weapons is publicly available. Our proposed algorithm is end-to-end trainable. A direction forecast module is prepared to foresee conceivable object direction for every area proposition. Our proposed model has done this task in two ways. The first is using angle of the object as classification and the second is using angle as a regression problem. We named our proposed method as Orientation Aware Weapons Detection (OAWD) because it provides not only detection but also the orientation of objects. 
     
For training and assessment purposes, a broad data set (OAWD) comprising of a wide assortment of guns and pistols is gathered from the web. It comprises photos of genuine scenes also ones from dramatizations and motion pictures. Figure 1 shows some sample random pictures of our data set.
\begin{figure*}[h]
    \centering \includegraphics[width=1\textwidth]{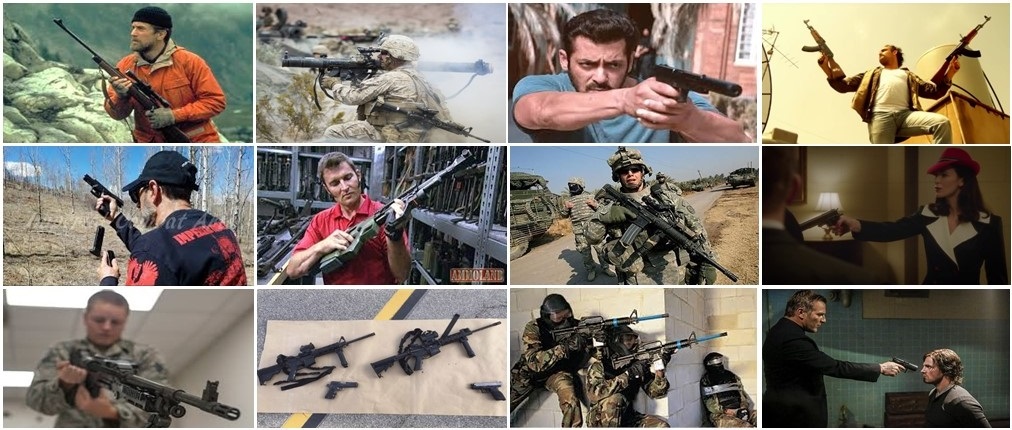}
    \caption{Illustrations from the proposed Orientation Aware Weapons Detection (OAWD) data-set }
    \label{fig:Sample_Picture_from_OAWD_Dataset}
\end{figure*}
This data set is annotated manually by using the roLabelImg tool which is an extended version of the LabelImg tool and provides an oriented bounding box. The data set, contains 6,400 pictures and has assorted attributes counting pictures catching different guns, various conditions, present varieties of people conveying weapons, what's more, pictures with guns without people. The proposed model is compared with FRCNN \cite{ren2015faster}trained on the same data set. Our work will be an impetus in making a difference out for oriented-aware weapons recognition. 
    
Our major contributions of this research work are listed underneath. 
\begin{itemize}
\item We present the first exhaustive work on guns location in RGB pictures. We analyse the drawbacks of using a horizontal bounding box for detection of oriented aware object detection e.g. pistols and guns by taking noise as background information. We propose orientation aware model in two ways: one is using angle as a classification problem and the second is using angle as a regression problem. Oriented bounding box has less background information as compared to axis-aligned windows. 
\item We also propose a first comprehensive oriented aware weapons dataset consisting of a total of 6400 images with single and numerous guns and pistols. Our dataset not only provides an oriented bounding box with angle information as a ground truth but also a provides horizontal bounding box as ground truth. We provide annotation of our dataset in three different formats for further research
\item Our proposed oriented aware model outperforms in results as compared to present state-of-the-art object detection algorithms e.g. Faster R-CNN.   
\end{itemize}
The remainder of this paper is sorted out as follows: Section 2 is about related work. Our proposed Orientation Aware Weapons Detection (OAWD) in visual data is described in Section 3. Section 4 is about the data set. Section 5 is about experiments done and analysis of results and Section 6 follows the conclusion and future directions. 

\section{Literature Review}
Research on visual gun recognition in pictures or recordings is very scanty and presently there is no committed gun indicator or gun benchmark data set for execution assessment. Almost 1000 object classes are present in the ImageNet data set\cite{deng2009imagenet} and just two are devoted to guns counting quickly discharging programmed weapons and gun or pistol with the spinning chamber, which is a very little assortment and not sufficient enough data set for detection purposes. Open Images data set V5 has also a handgun class but it has only around 600 images of this which are not enough. Moreover, the orientation of these data set is horizontal, not oriented box. Olmos et al. has applied Faster RCNN \cite{ren2015faster}for detection of a handgun in recordings\cite{olmos2018automatic}, while no outcomes have been accounted for on rifle identification.  Akcay et al. has applied RCNN\cite{girshick2014rich}, FRCNN\cite{ren2015faster}, Yolo V2\cite{redmon2017yolo9000} and RFCN\cite{dai2016r} for object recognition inside x-ray stuff security symbolism\cite{akcay2018using}. As opposed to these works, we first time, address the issue of oriented aware visual gun and riffle recognition in pictures in a more comprehended way.
\begin{figure}[ht]
	\centering
		\includegraphics[scale=1]{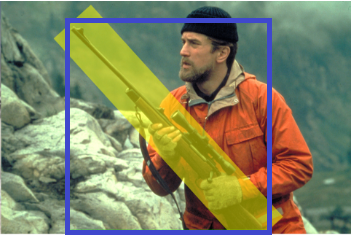}
	\caption{An example iamge from OAWD dataset; Illustrating that horizontal bounding box has a lot of irrelevant information in background while oriented bounding box (highlighted in yellow color) detected has successfully reduce this issue. }
	\label{FIG:1}
\end{figure}

Despite the fact that very limited research work has been done for oriented-aware weapon detection, comprehensive research work is available for creating generic object recognition algorithms. Tao Gong et al. \cite{gong2019using} proposed a unique end-to-end framework that improves object detection by using multi-label classification as an assistant assignment. But they didn't provide any method for oriented aware object detection and our proposed method provides orientation aware object detection. Y.Xiao et. all presented a review analysis of different deep learning models for object detection in detail. They also presented a brief overview of different traditional object detection methods. YOLO family \cite{redmon2016you,redmon2017yolo9000,redmon2018yolov3} which are based on deep convolutional neural network are very good detectors for real-time object detection. YOLO\cite{redmon2016you} is very speedy and it can be applied in real-time detection. It provides more localization errors than other detector methods. YOLO v2 \cite{redmon2017yolo9000}and v3\cite{redmon2018yolov3} gives good results than YOLO \cite{redmon2016you} because it uses multi-scale training, batch normalization and  anchor boxes. However, the performance of YOLO family is lower than Faster RCNN\cite{ren2015faster}. YOLO family provides horizontal bounding boxes but our proposed method provides an oriented bounding box method.  
    
As compared to YOLO\cite{redmon2016you}, SSD\cite{liu2016ssd} is also a single-stage object recognition algorithm. VGG16\cite{han2015deep} network is used as base architecture and instead to classify from the last fully connected layer of VGG16 they add four extra convolution layers using smaller filters and take samples from multiple feature layers and then combine them in output feature space. Performance of SSD\cite{liu2016ssd} compromises if objects are small. Another variation of SSD\cite{liu2016ssd} is DSSD \cite{fu2017dssd} as they used ResNet-101 architecture as base network instead of VGG16 which is used in SSD\cite{liu2016ssd}. DSSD\cite{fu2017dssd} uses deconvolutional layers that use skip connections and give rich feature map for better recognition. A recent study shows that the performance of DSSD\cite{fu2017dssd} is less as compared to Faster RCNN\cite{ren2015faster}. Also, because of deconvolutional layers, the computational complexity of DSSD\cite{fu2017dssd} is higher. SSD and DSSD also provide axis-aligned bounding box for object detection. They do not have ability to provide an oriented bounding box for object detection. SSD and DSSD both models do not incorporate angle in any way for orientation. Our proposed model incorporates angle and provides an oriented aware bounding box.  
    
In weapon detection issues, weapons may show up in very little size contrasted with the general picture measurements. In recognition of small objects, picture size, scale, and relevant data may have essentialness for learning profound model \cite{hu2017finding}. M.Grega et al. \cite{grega2016automated}proposed a model for automated recognition of knives and firearms in CCTV video. They used a conventional approach for the recognition of firearms. They only focused on pistol detection in weapons. Although they detect pistol but detection is not oriented aware while our proposed model provides oriented bounding box for object detection.  T.Y.Lin et al. proposed novel focal loss to handle the problem of class imbalance \cite{lin2017focal}. R. Olmos et al. proposed a model based on Faster RCNN\cite{ren2015faster} which generates an automatic alarm if the model detects a handgun in video\cite{olmos2018automatic}. They just focused on handgun detection\cite{olmos2018automatic}. They do not focus that at which angle handgun is present. Our proposed model also tells this that at which angle the handgun is present. R.K Tiwari et al. \cite{tiwari2015computer} proposed a conventional based approach for gun recognition in visual data with the help of Harris interest point detector\cite{harris1988combined} and FREAK\cite{alahi2012freak}. Their performance does not perform well on illumination change. Chen et al. and Liu et al. have stressed the essentialness of context and example relationship for precise detection of object\cite{liu2018structure}, \cite{chen2018context}. For instance, in the event that an individual is holding a tennis racket, at that point there ought to be a ball close by. Anyway, on account of weapons, many of the logic may remain unrelated contextually to the existence or non-existence of weapons. 
    
Huang and Rather et al. have performed an analysis in detail between accuracy-speed trade off\cite{huang2017speed}, in which they showed that Faster RCNN \cite{ren2015faster}is more steady contrasted with the other locators. Faster RCNN \cite{ren2015faster}has developed to the current structure in the wake of experiencing numerous varieties. In the past adaptations, RCNN was using Selective Search method \cite{uijlings2013selective}for generating region proposals, applied profound convolutional systems on each proposal to remove low-level features, and then SVM \cite{furey2000support}was used for classification. Later on, the RPN module introduced in Faster RCNN \cite{ren2015faster}that share convolutional features of the full image with the detection network. They used the sliding window method on a convolutional feature map to generate region proposals. ROI pooling layer was also introduced for obtaining fixed-size regions of interest. At every position of the feature map, they yield a total of 9 anchors using three scales and three ratios. They performed NMS to obtain high probability region proposals. After applying ROI pooling on selected region proposals, the classifier is applied. Presently an enormous number of FRCNN have been proposed, particularly FPN and Mask RCNN\cite{lin2017feature},\cite{he2017mask} are more significant. All these methods give a horizontal bounding box at the output. Our proposed model provides an oriented aware model not only on the basis of angle classification but also on the basis of angle regression. According to our best knowledge no one has done orientation using angle regression. Thus our proposed model Orientation Aware Weapon Detection (OAWD) is unique and different from present object detection algorithms.  

According to our best knowledge, no comprehensive public dataset is available related to weapons that provide oriented bounding box as ground truth. Although it can be seen in Table 1 that some public available datasets (Open Images Dataset V5\cite{kuznetsova2020open} , IMAGENET Dataset \cite{fei2010imagenet}, Handgun Dataset \cite{olmos2018automatic}, Guns Dataset \cite{gunsdataset} and Weapons Detection Dataset \cite{olmos2018automatic}) have weapons images but they provide horizontal bounding box as ground truth not oriented bounding box. Moreover, they have very less weapon images. Although HRSC2016 \cite{liu2016ship} dataset and DOTA dataset \cite{xia2018dota} provide oriented bounding box but they have no weapon class and weapon images. In contrast to other publicly available datasets which we mentioned above, our dataset has a total of 6400 weapon images of two classes: one is a gun and the second is a pistol. Our dataset provides both oriented bounding box with angle information and horizontal bounding boxes as ground truth. We are also working to expand this dataset further.
\section{Dataset}
For automatic detection of weapons, no oriented aware public data set is available for guns and pistols for training. In this research work, we propose the first oriented aware data set for weapons which we named "OAWD data set".
\begin{figure}[ht]
	\centering
		\includegraphics[scale=0.40]{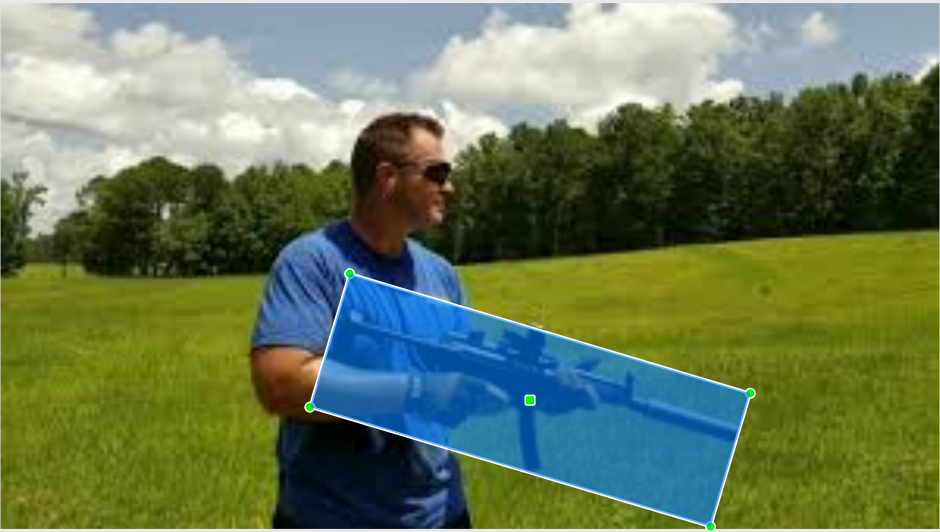}
	\caption{An example image from OAWD dataset showing manual annotation for Ground preparation}
	\label{FIG:2}
\end{figure}
\subsection{OAWD Data Set}
In present years, Datasets have assumed a significant part in the information-driven examination. Enormous datasets like MSCOCO are instrumental in advancing object detection and captioning. With regards to the classification undertaking and scene acknowledgment task, the equivalent is valid for ImageNet and Places, separately. Moreover, in oriented object detection, a dataset looking like MSCOCO and ImageNet both as far as quantity of images and oriented annotations have been missing, which gets one of the fundamental obstructions for research especially in detecting oriented aware weapons detection. Accordingly, a comprehensive large-scale and challenging-oriented aware weapons data set is basic for advancing research in this domain.
\begin{table}
\centering
\caption{Comparison of OAWD dataset with other publicly available dataset which has weapons images }
\begin{tabular}
{| p{0.4\textwidth} | p{0.2\textwidth} | p{0.12\textwidth} | p{0.12\textwidth}|}
\hline
\textbf{Dataset} & \textbf{Annotation Method} & \textbf{Weapon class} & \textbf{Weapon Images} \\ \hline
HRSC2016 \cite{liu2016ship} & Oriented BB & No & 0 \\ 
DOTA \cite{xia2018dota}&  Oriented BB & No & 0  \\ 
Open Images Dataset V5 \cite{kuznetsova2020open} &  Horizontal BB & yes & 650  \\ 
IMAGENET Dataset \cite{fei2010imagenet}&  Horizontal BB & yes & 1200  \\ 
HandGun Dataset \cite{olmos2018automatic}&  Horizontal BB & yes & 795  \\ 
Guns Dataset \cite{gunsdataset} &  Horizontal BB & yes & 333  \\ 
Weapons Detection Dataset \cite{olmos2018automatic}&  Horizontal BB & yes & 3304  \\ \hline
OAWD Dataset&  Oriented BB & yes & 6400  \\ \hline
\end{tabular}
\end{table}

We contend that a comprehensive weapons dataset ought to have four properties, specifically, 1) an enormous number of pictures, 2) numerous examples per classifications, 3) appropriately oriented bounding box along with the angle, and 4) different classes of weapons, which make it way to deal with real-time applications. While presently available dataset has lack of comprehensive dataset of weapons, even some datasets do not have weapon class. With regards to general object detection datasets, ImageNet dataset \cite{fei2010imagenet} and MSCOCO \cite{lin2014microsoft} are supported by analysts because of the enormous number of pictures, numerous classifications categories, and good annotations. ImageNet is the largest dataset as compared to other datasets as it has the biggest number of pictures. Notwithstanding, the average number of objects per picture is less than MSCOCO and our OAWD, in addition to the constraints of its spotless foundations and deliberately chosen scenes.

\subsection{Annotations of OAWD dataset}
\subsubsection{Images Collection}
We have gathered our own data set using web scrapping with different keywords e,g, gun, rifle, pistol, weapon, hand gun, gun with human and rifle in the human hand, etc. Then manually delete those images which were not related to guns or rifles. We have also collected images using the google chrome extension "Download All Images" to download all images of a specific page. We have also downloaded videos of movies from YouTube and then extract images from those videos which have pistols or guns in the image. We have also downloaded videos of robbery in the shop and such videos are very low resolution and images gathered from these videos have very low resolution. The purpose of gathering low-resolution images was that so our model can detect real-time weapons. Because we knew that normally video quality of cameras installed in shops are of very low resolution. 
\begin{table}
\centering
\caption{Statistics of Data set after splitting into training and testing data set}
\label{table:schriber2018}
\begin{tabular}{| l | l | l | l | l|}
\hline
& \multicolumn{4}{c}{\textbf{Weapon Count }} \vline \\\cline{3-5}
\textbf{Dataset} & \textbf{Total Images} & \textbf{Gun} & \textbf{Pistol} & \textbf{Total weapons}\ \\ \hline
Training & 5149 & 4341 & 3206 & 7547\\ \hline
Test &  1249 & 642 & 825 & 1467  \\ \hline
\end{tabular}
\end{table}

\subsubsection{Class Selection}
Two classes are chosen and manually annotated in OAWD dataset, including gun and pistol. Gun can be of any category e.g. bolt action riffles Remington 700 and Howa 1500, lever action riffles Winchester 94 and Marlin 336 and semi-automatic riffles AR-15 and Browning BAR while pistol is shorthand gun includes Moss-berg 500, Remington 870, Smith \& Wesson Model 686, Ruger GP100 and Glock 17, etc. 

\subsubsection{Annotation Method}
We think about various methods of annotation. In computer vision, numerous visual ideas, for example, attributes, objects, relationships, and region descriptions are annotated using bounding boxes. A typical depiction of bounding boxes is ($x_c$, $y_c$,w and h), where ($x_c$, $y_c$) is the middle location of rectangle and w, h are the width and height of the bounding box. After gathering images, the next part was to annotate this data using such a tool that gives rotated annotation. We have used roLabelImg~\cite{rolabelimg} tool for annotation of images. roLabelImg tool is an extended version of LabelImg which provides annotation in rotated format. This tool provided annotation in XML format like PASCAL VOC format. The annotation format of our data set is Xc, Yc, width, height, and angle. Figure 4 shows sample ground truth of an image annotated by the roLabelImg tool.
\subsubsection{Dataset Splits}
Table 1 shows the statistical data set of training and testing after randomly splitting data set into training and testing with the ratio of 80\% vs 20\%. It also shows that how many objects of the Gun and Pistol class are present in both the training and testing data set. It can also be seen in table 2  that how many total objects are present in the training and testing data set. 
\subsection{Properties of OAWD}
\subsubsection{Image Size}
The spatial dimension of images in the OAWD dataset is not fixed. The minimum spatial size of a single image in our dataset is 104 x 104 while the maximum dimension of a single image in our dataset is 6720 x 4480. We did manual annotations on the full image.
\subsubsection{Annotation Formats of OAWD Dataset}
Annotation formats are very important for fast research. OAWD dataset is available in the following three different formats:
\begin{itemize}
\item CNTK Faster RCNN Format
\item Tensor Flow Pascal VOC Format
\item YOLO aware Format
\end{itemize}
\subsubsection{Angle Wise Distribution of OAWD dataset}
We have divided our angle into 8 classes starting from class 1 to class 8 for analysis of data more statistically.  Range of class 1 is from $-11.25^0$ to $11.25^0$, class 2 is from $11.25^0$ to $33.75^0$, class 3 is from $33.75^0$ to  $56.25^0$, class 4 is from $56.25^0$ to $78.75^0$, class 5 is from $78.75^0$ to $101.25^0$, class 6 is from $101.25^0$ to  $123.75^0$, class 7 is from $123.75^0$ to $146.25^0$, class 8 is from $146.25^0$ to  $168.75^0$.
\begin{figure}[ht]
	\centering
		\includegraphics[scale=.30]{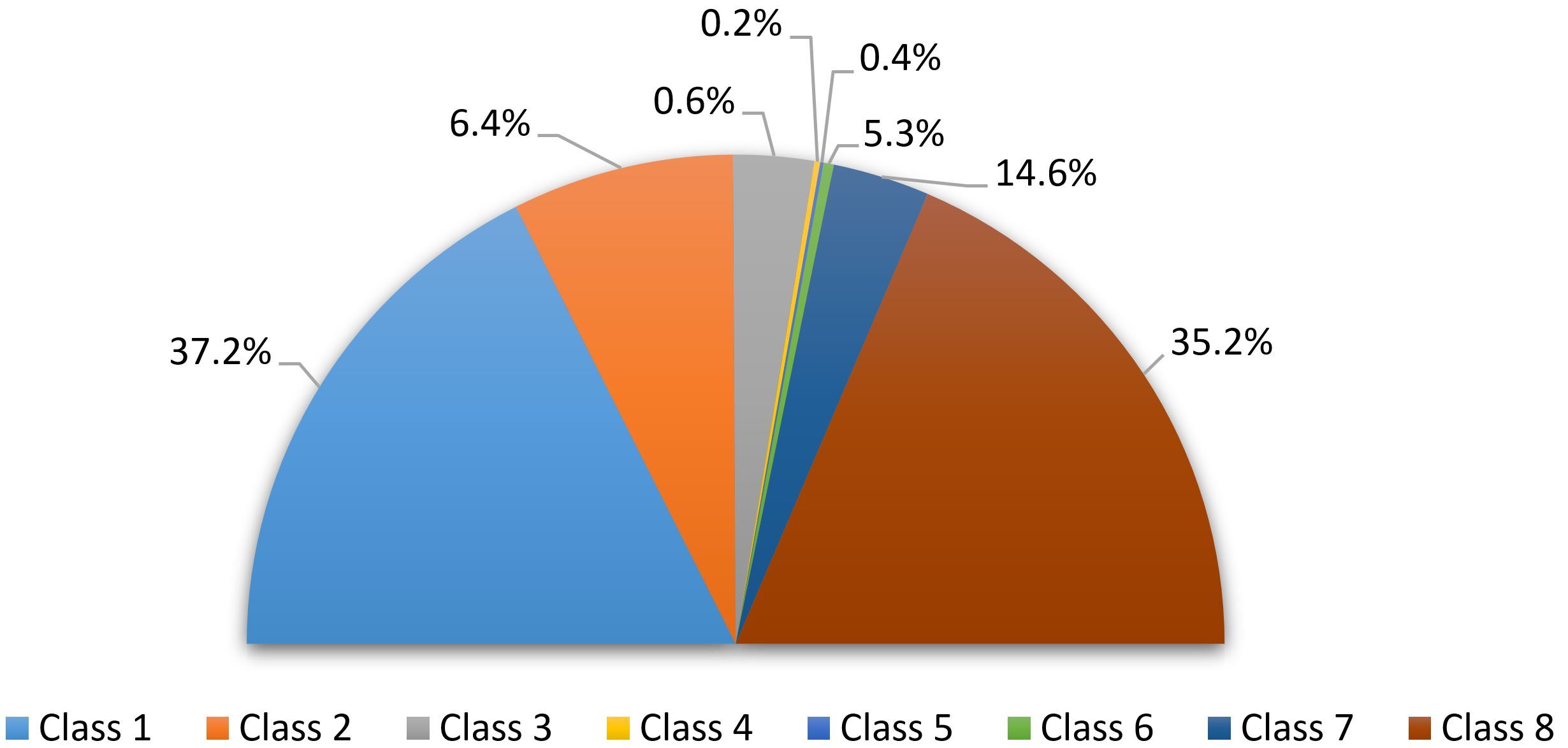}
	\caption{Data set distribution Pie Chart according to 8 classes of orientation angle}
	\label{FIG:3}
\end{figure}
Figure 4 shows data more statistically according to 8 classes of angle. Figure 5 shows that how many weapons are in every class. It is clear from figure 5 that class 1 has 2654 objects while class 8 according to angle has the most number of objects 2804 and class 5 according to angle has 17, the least number of objects. Minimum 1 weapon and a maximum of 10 weapons are present per image while on average there are 1.40 weapons per image in our data set. Class 4,5 and 6 has very few number of weapons because class 5 belongs to angle 90 while class 4 and class 6 belongs to around 90 degrees and it is a very rare case in real life that a man will have a gun in hand at 90 degrees because at 90-degree gun or riffle nozzle will have a direction towards the sky. Moreover, it is not dangerous for the next person if a man has a weapon in hand at 90 degrees and it is not easy for a man to pick a gun in hand at this angle for a long time.                
\begin{figure}[ht]
	\centering
		\includegraphics[scale=.30]{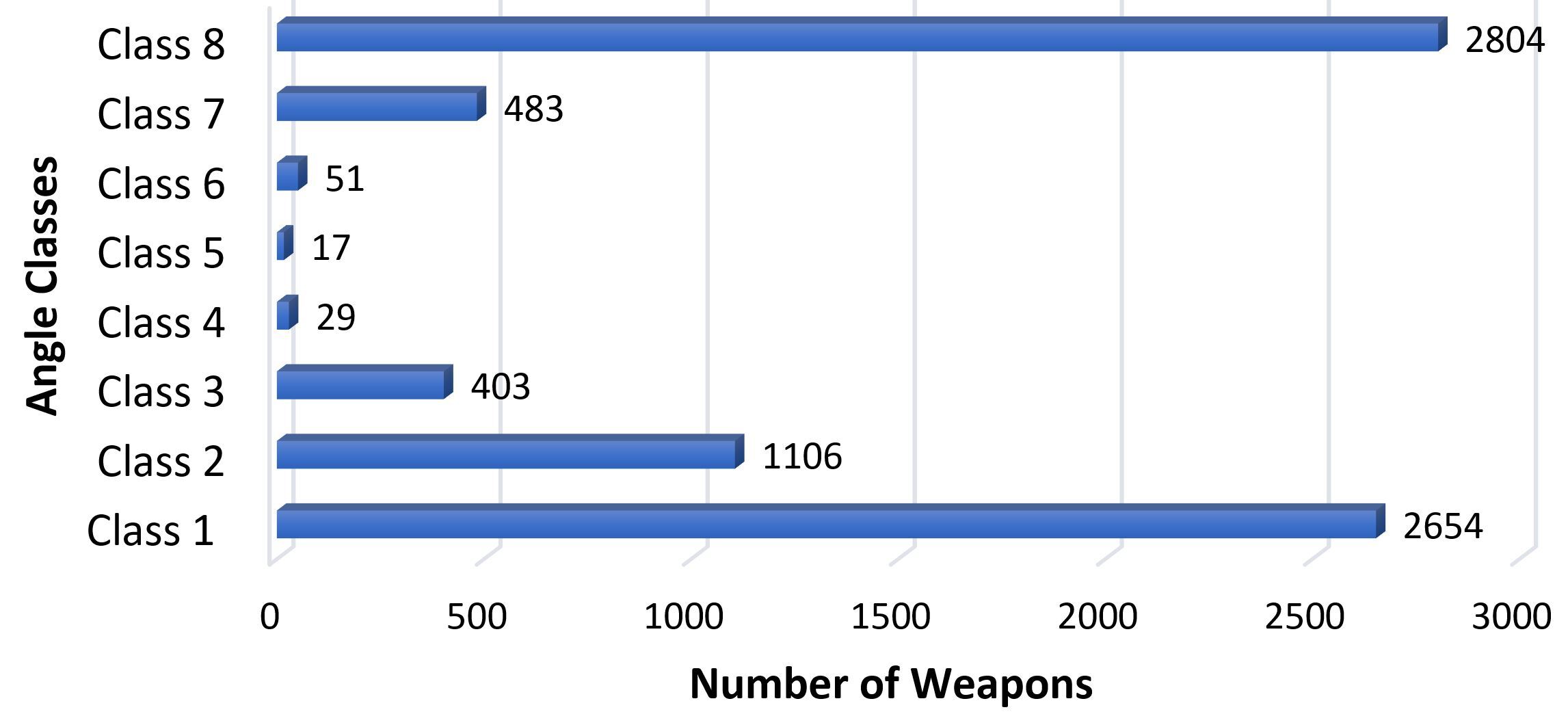}
	\caption{Dataset distribution bar Chart according to 8 classes of orientation angle}
	\label{FIG:e}
\end{figure}

\section{Proposed Model}
Most of the present object detection algorithms provide horizontal bounding boxes which contain a lot of background information and the object is not more localized. To conquer this issue, we propose Orientation Aware Weapons Detection (OAWD) algorithm which provides orientation in two different ways. In the first way, the model provides orientation using angle as classification and in the second way, it gives orientation using angle as regression. In first way, we can only tell the specific angle of the object but in a second way we can tell any angle of the object. According to our best knowledge no one has done orientation by taking angle as regression. The following two ways are more described in detail below here.
\subsection{Angle Classification Method}
OAWD using angle classification mainly comprises of computing deep feature map using ResNet, RPN, RoI pooling followed by FC (fully connected) layers. For this purpose, we also prepared our data set for training and define the angle class. Only at inference time we apply linear rotation transformation for obtaining rotated bounding box. Every one of these segments is clarified in the accompanying subsections. Figure 6 shows the architecture diagram of OAWD using angle classification. 
    
\begin{enumerate}
\item Dataset Preparation: We know Faster R-CNN takes text document as input but in our data set annotation was in XML format. So we first converted that XML sheet into an excel sheet. Now in our excel sheet we have the following information of every image as input to Faster R-CNN; image name, image width, height, X1, Y1, X2, Y2, object class, and angle class.
\begin{figure*}
\includegraphics[width=1\textwidth]{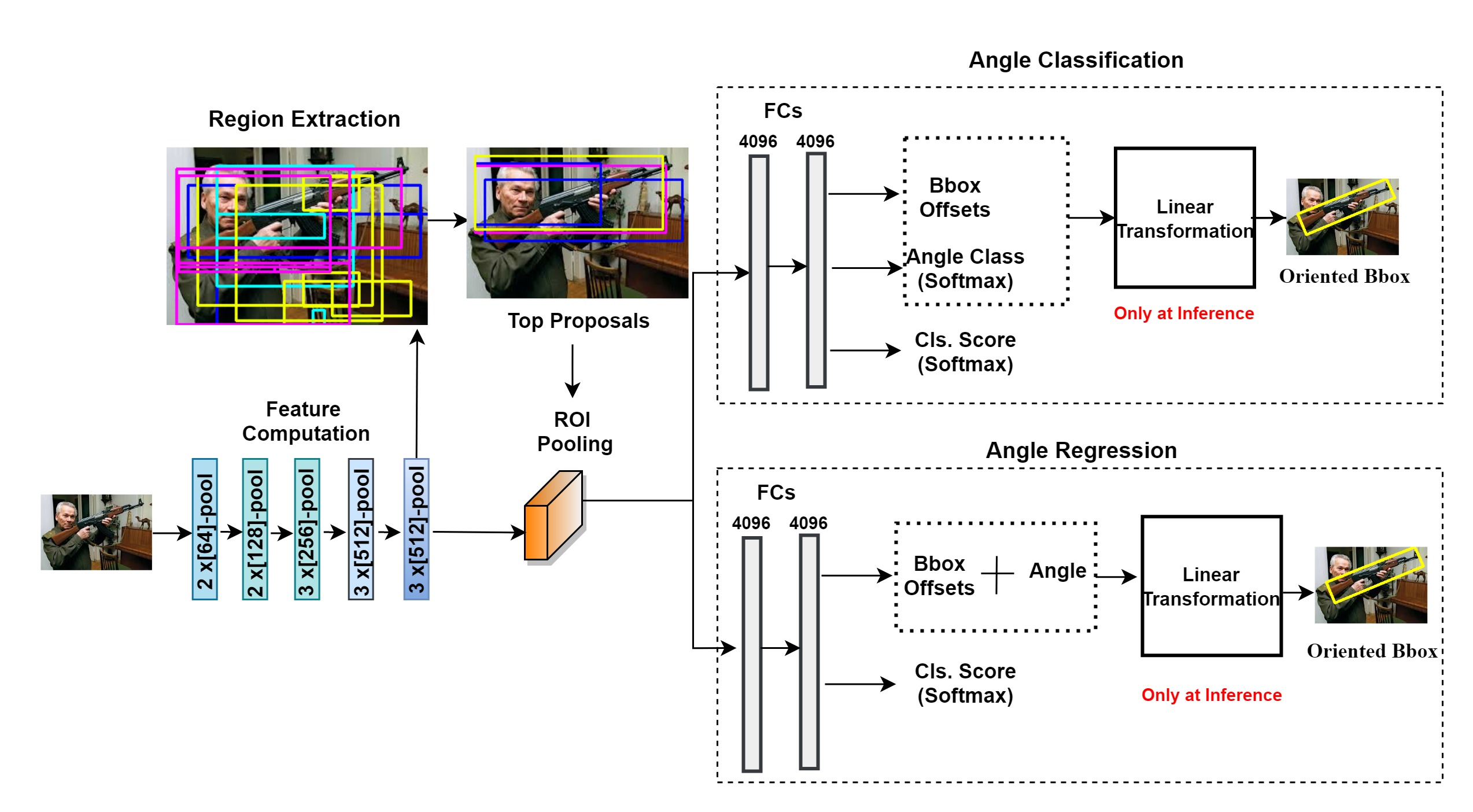}
\caption{Our proposed OAWD model architecture diagram with details of feature extraction, selection of top proposals, RoI pooling and detection. It is shown that our proposed model provides orientation in two ways one is using angle classification and other is using angle regression}
\label{fig:Network_Architecture_Image}
\end{figure*}   
            
\item Define Angle Class: In XML format of data set we have angle information in radian which is in degree from $0^0$ to $180^0$. when we convert this XML file into an excel sheet we converted that angle into the following 8 classes; class zero is from $-10^0$ to $12.5^0$, class 1 is from $12.5^0$ to $35^0$, class 2 is from $35^0$ to $57.5^0$, class 3 is from $57.5^0$ to  $80^0$, class 4 is from $80^0$ to $102.5^0$, class 5 is from $102.5^0$ to $125^0$, class 6 is from $125^0$ to  $147.5^0$ and class 7 is from $147.5^0$ to  $170^0$. So now we have additional information of angle in input.
\item Computing Feature Map: After defining angle class and dataset preparation, We used VGG like architecture for feature extraction from an input image. Though any deep neural network can be used for feature extraction. Here we can use any CNN network for purpose of feature extraction.
Convolution operation can be represented as shown in equation 1 given below:
\begin{equation}
F_n^m(a,b) = {\sum_{j}\sum_{u,v} i_j(u,v).e_n^m(c,d)}
\end{equation}

in which $F_n^m$(a,b) shows feature map element of $m^{t^h}$ kernel of $n^{t^h}$ layer, j represents channel, $i_j$(u,v) represents element of an image with respect to $j^{t^h}$ channel and $e_n^m$(c,d) represents element of $m^{t^h}$ kernel with respect to $l^{t^h}$ layer. Because input images may have differ in size, that's why feature map may also differ in spatial dimensions but depth of feature map (512) will remain same. The feature map weights can be learned using following equation 2: 
\begin{equation}
W_{}(_j{}_,{}_c{}_) = G(x_{}j_,{}_c{},w)
\end{equation}
in which G represents activation function which is ReLU in our case, $x_j$ represents last layer output, w represents weights of $1^{s^t}$ convolution layer in OAWD and $W_{}(_j{}_c{}_)$ represents newly learned weights of feature map at position j with respect to c channel.  
\item RPN (Region Proposal Network): RPN network takes feature map as input. RPN is arbitrarily initialized and afterward trained on the training dataset of OAWD dataset to create objectness score and proposals of an object exist in a picture. At every location of the feature map total of 9 anchor boxes using three different scales and three ratios are used to handle the different sizes of objects. RPN module generates almost 2000 region proposals. To additionally decrease the computational complexity, just a small amount of these best proposals is chosen for additional handling which is then used as input to the RoI pooling. We selected the top 300 region proposals on the basis of overlap between generated region proposals and the ground truth box. If the overlap between generated region proposals and ground truth box is greater than 0.7 we selected that box and discard all other remaining region proposals. If we have more than 300 region proposals that have greater than 0.7 overlap with the ground truth box then we select only the top 300 proposals which have greater overlap with ground truth box. We labeled these region proposals as given in equation 3:  
\begin{equation}
    RPN Label=\begin{cases}
    Gun,  \text{if} IOU_1\geq0.7.\\
    Pistol, & \text{if $IOU_2>=0.7$}.\\
    Background, & \text{Otherwise}.
    \end{cases}
\end{equation}
where $IOU_1$ is the IOU between ground truth bounding box of gun and RPN box and $IOU_2$ is IOU between ground truth bounding box of pistol and RPN box. The regression loss function for the bounding box is given in equation 4: 
\begin{equation}
L_r{}_e{}_g{}_r{}_e{}_s{}_s{}_i{}_o{}_n{}(t^l,v) = {\sum_{j\in({x,y,w,h})} Smooth L_1{}(t_{}j^l - v_j)}
\end{equation}
in which $t^l$ represents bounding box offsets for l object classes and v is ground truth bounding box offsets and $SmoothL_1$ can be calculated using equation 5: 
\begin{equation}
SmoothL_1 (y) =\begin{cases}
    0.5y^2, & \text{if $|y|<1$}.\\
    |y|- 0.5, & \text{Otherwise}.
    \end{cases}
\end{equation}

\item Classification Layers: After getting the top 300 region proposals we do ROI pooling before applying fully connected layers. Best recommendations of proposals from RPN are used as input to ROI pooling and it chooses the corresponding feature map which we have already computed by using ResNet. Since fully connected layers only take fixed-size input which is acquired by applying ROI pooling on the feature map. Then after ROI, we apply 2 fully connected layers and at the output it gives bounding box offset, object class score, and angle class score. We used Softmax loss for angle classification as well. The loss function we used is defined as: 
\begin{equation}
L_c{}_l{}_s{}(a)_i = \frac{e^{a_m}}{\sum_{n=1}^{K} e^{a_n}}
\end{equation}

where a is input vector which is equal to (a1,a2,....,aK), m,i=1,2,3,....,K and K is the number of classes which are gun and pistol in our case.  Now at the output, we have types of information; one is bounding box offset, second is object class and third is angle class. We used RELU activation to introduce non-linearity in our model. The activation function can be defined as: 
\begin{equation}
O_n^m = {G_a(f_n^m)}
\end{equation}
in which $G_a$ is activation function which takes output of convolution that is $f_n^m$ and $O_n^m$ is output for $n^{t^h}$ layer. 
\item Linear Transformation: After getting output we apply rotation transformation on 4 corner points of the bounding box. Linear transformation module takes bounding box coordinates and angle as input and it gives rotated points of bounding box according to angle. Then we draw the bounding box according to the new bounding box corner points and finally, we get the rotated bounding box. For example, if we have any point a,b in 2-D. we can rotate that point around a specific position by using equations 8 and 9. These equations give new values of a,b of a point after rotation using given any angle. 
\begin{equation}
    a' = a*cos\theta - b*sin\theta
\end{equation}
\begin{equation}
    b' = b*cos\theta + a*sin\theta
\end{equation}
\end{enumerate}
    
We have divided our angle into 8 classes as describes above. But during rotation transformation, we used 8 separate angles for every class. For class 1 we choose angle zero although class zero has an angle range between $-10^0$ to $12.5^0$. Similarly for class 2,3,4,5,6,7 and 8 we choose angle $22.5^0$, $45^0$, $67.5^0$, $90^0$, $112.5^0$, $135^0$ and $157.5^0$ respectively. 
\subsection{Angle Regression Method}
Since problem was to detect and classify weapons along with their orientation. In the previous part, we deal with angle as a classification problem but it does not predict all angles from $0^0$ to $180^0$ and we can not tell the exact angle of an object present in an image. So here in this part, we deal with angle as a regression problem so that we can predict all angles from $0^0$ to $180^0$ rather than some specific angles. According to our best knowledge, there is no method or model available right now which deals with angle as a regression problem. There is only a change in data set preparation for training and in classifier layers at the end of the model. All other work is the same as it was in the classification part. For data set preparation, we first converted that XML sheet into an excel sheet. Now in our excel sheet, we have the following information of every image as input to Faster R-CNN; image name, image width, height, X1, Y1, X2, Y2, object class, and angle class. In the XML format of data set, we have angle information from $0^0$ to $180^0$.For this purpose when we converted XML file of every image into an excel sheet, we did not divide an angle into classes. We took the angle as it is as it was in the XML file. Angle information in the XML file is in radian. 
            
We took an angle in such a way that angle information is in the range of 0 to 1 by dividing every angle into 8. Since now the angle is in the range of 0 to 1, so it is easy to do a regression of angle. In classification layers, we use the smooth L1 loss for regression of angle. The model gives two outputs in this case. One tells about object class and the second output tells not only about bounding box offsets but also gives one additional offset value which is for angle. Then using the rotation transformation describes above, we get a rotated bounding box. 

\section{Experiments and Results}
We compare the results of the proposed OAWD model architecture with Faster RCNN which provides a horizontal bounding box with good detection results. Our model not only provides good results than Faster RCNN but also provides an oriented bounding box at any angle. 
\subsection{Training Settings}
Training has been done on GPU and we used NVIDIA Ge Force 1080 TI GPU for training. We used a system that has 64GB RAM, 1TB hard disk along with 250GB SSD. The total number of epochs during training are 200 and one epoch length is 1000. RPN overlap is 0.7 and IOU is set to 0.5 during non-maximum suppression. The learning rate is set to 1e-5.
\subsection{Results \& Analysis}
We used mAP (mean average precision) for the evaluation of our proposed model. We assess the IOU (Intersection over Union) of the identified and the ground truth bounding boxes. If the IOU of an instance is greater than or equal to 0.5, we considered it as TP (True Positive), else it is FP (False Positive). Precision is calculated as TP/(TP+FP). At each level, average precision is calculated and. A similar procedure is repeated for every class independently and an average of all classes is presented as mAP.  
\begin{table}
    \centering
    \caption{Quantitative Results of OAWD using angle classification and angle regression and compared with Faster RCNN which gives horizontal bounding box}
    \label{table:12}

    \begin{tabular}{|  p{0.42\textwidth} | l | l |}
                    \hline
                    & \multicolumn{2}{c}{\textbf{OAWD Dataset }} \vline \\\cline{2-3}
                    \textbf{Network} & \textbf{Training (map)} & \textbf{Testing (map)} \\ \hline
                    Faster R-CNN & 86.80 & 81.2 \\ \hline
                    Proposed Model OAWD Classification &  88 & 82.90  \\ \hline
                    Proposed Model OAWD Regression &  87 & 81  \\ \hline
                    \end{tabular}

\end{table}
                
Table 3 shows that the map of training and testing of OAWD using the angle classification approach is greater than Faster R-CNN. It is because our data set is oriented and it takes less background information than the horizontal bounding box data set which Faster R-CNN takes. 
\begin{figure*}
            \centering       
             \includegraphics[scale=0.40]{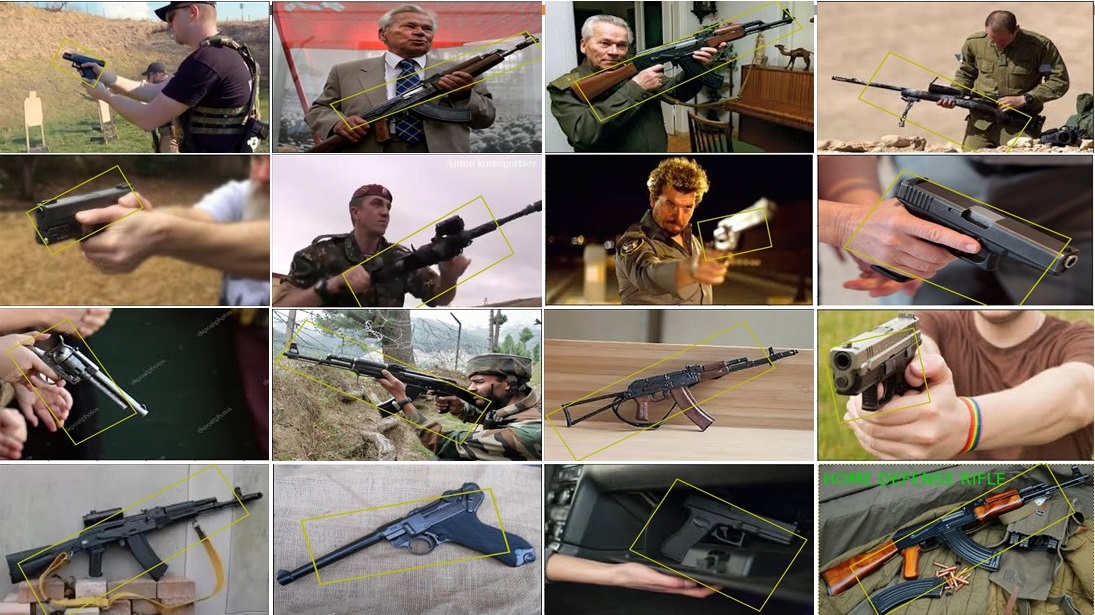}
                \caption{The prediction of OWAD model architecture: The bounding boxes are shown as overlapped on the original images of the proposed OWAD dataset}
                \label{fig:Foreground_Encoded_Image}
\end{figure*} 
Figure 7 shows the prediction results in form of bounding boxes overlapped on the original images in the OWAD dataset. 
Table 3 also shows that although mAP of OAWD using angle regression is higher than Faster R-CNN but slightly less than OAWD using classification.

\begin{figure*}
	\centering
	\includegraphics[scale=0.45]{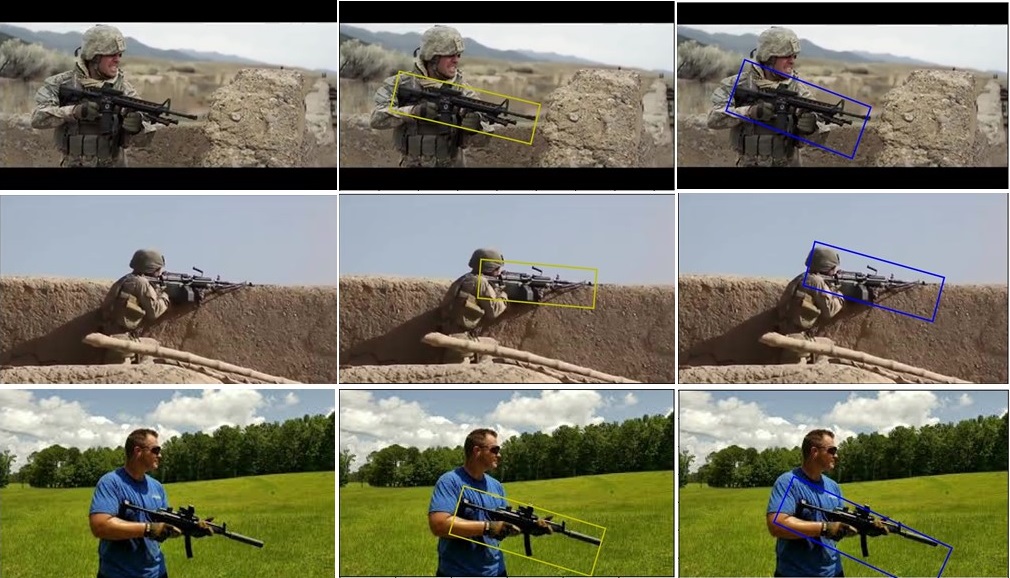}
	\caption{Angle Regression based Predictions of Oriented Weapons Detection: left most image is the original image, center image is showing ground truth box and right most image shows predicted bounding box.}
	\label{FIG:4}
\end{figure*}

Figure 8 shows qualitative results of OAWD using angle regression. Qualitative results are somehow very close to OAWD (classification) results. Because we have done regression of angle in this model and it can be seen in figure 9 that due to slight change in angle our bounding box rotates according to that. We also have presented an analysis of the results of OAWD (classification) and OAWD (regression) with Faster R-CNN. We have computed mAP at different IOU ratios. We have computed both training and testing mAP at 0.25,0.5 and 0.75 IOU ratio and shown results in form of a table which can be seen in table 4.

\begin{table}{}
\centering
\caption{Quantitative results at different IOU}
\label{table:schriber2019}

\begin{tabular}{|  p{0.42\textwidth}  | l | l | l |}
\hline
& \multicolumn{3}{c}{\textbf{OAWD Dataset }} \vline \\\cline{2-4}
\textbf{Network} &  \textbf{Testing }& \textbf{Testing} & \textbf{Testing} \\ \hline
\textbf{}  & \textbf{map@0.25} & \textbf{map@0.5} & \textbf{map@0.75} \\ \hline
Faster R-CNN  & 86.8 & 81.2  & 58.50 \\ \hline
Proposed Model OAWD Classification & 87.20  & 82.90 & 60.5 \\ \hline
Proposed Model OAWD Regression & 88 & 81 & 59.5 \\ \hline
\end{tabular}

\end{table}

Table 4 shows that testing map of OAWD using regression at 0.25 is higher than other two models. While testing map of OAWD using classification is higher than other 2 models.  

\begin{figure*}
\centering
\includegraphics[scale=0.40]{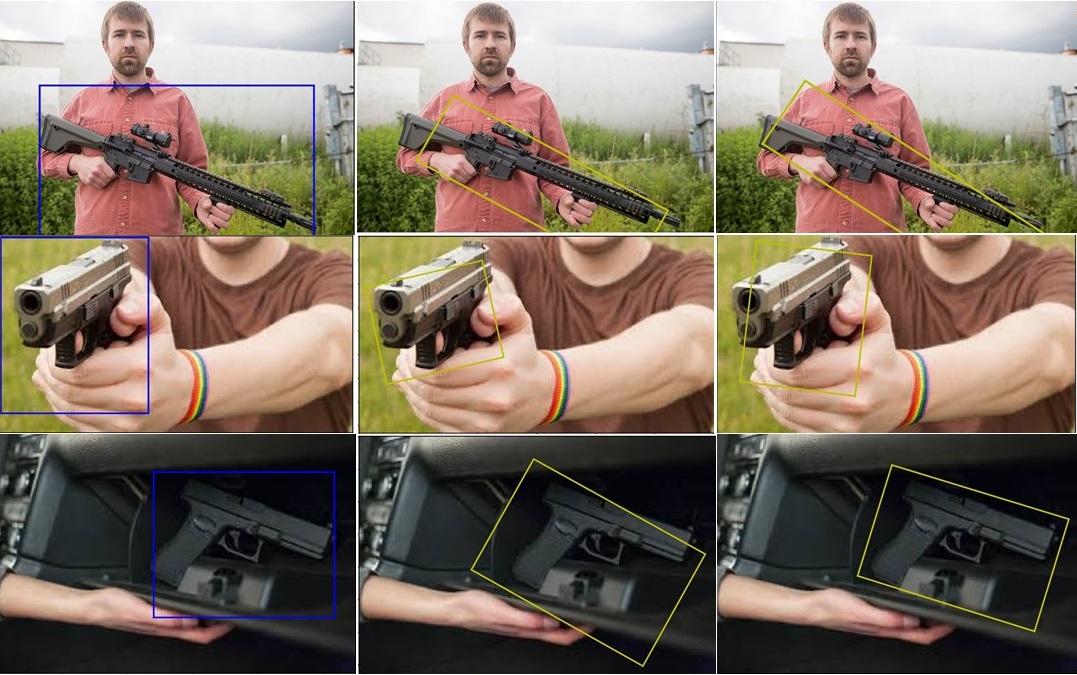}
\caption{Comparison of Predicted Bounding Boxes for Weapon's Detection; (Left most)Faster R-CNN; (center) Proposed OWAD Model using classification; (Right most)Proposed OWAD Model using angle regression}
\label{fig:Foreground_Encoded_Image1}
\end{figure*} 

Figure 9 shows the comparison of qualitative results of three models starting from left to right Faster R-CNN, OAWD using angle classification, and OAWD using angle regression. The first 2 pictures have shown good results on both oriented models while OAWD suing regression model gives a slightly bad result on the third picture. 
From the results and comparison given above it can be seen that the oriented model gives better results than the horizontal model. In the oriented model, we can easily detect only objects and it gives less background information than the horizontal model. The oriented model focuses more on objects or we can see oriented gives higher foreground to background ratio.
\subsection{Ablation Study}
We have tested our model using different parameters. First, we tried to train our model on 4GB GPU memory but our training was not giving good results. It is necessary to have a good GPU of up to 11 GB memory for the training model. Below here we have discussed at which parameters we have checked our model step by step. 
\begin{enumerate}
\item Using ResNet as backbone network: We also used Res-Net as backbone network in our architecture for feature extraction and compare results with the base model. It can be seen in table 5 that using ResNet as a backbone does not give good results. Although ResNet architecture is better than VGG but here we got better results using VGG as backbone. The reason is that when the dataset is simple and not complex and small dataset then VGG gives better results. 
\begin{table}
\centering
\caption{Comparison of quantitative results using VGG and ResNet as backbone network}
    \label{table:8}

    \begin{tabular}{|  p{0.42\textwidth}  | l | l |}
                    \hline
                    & \multicolumn{2}{c}{\textbf{OAWD Dataset }} \vline \\\cline{2-3}
                    \textbf{Network} & \textbf{Backbone Network } & \textbf{Testing (map)} \\ \hline
                    Faster R-CNN & VGG16 & 72.98 \\ \hline
                    Proposed Model OAWD Classification &  VGG16 & 82.90  \\ \hline
                    Proposed Model OAWD Regression &  VGG16 & 81  \\ \hline
                     Proposed Model OAWD Classification &  ResNet & 73.70  \\ \hline
                    Proposed Model OAWD Regression &  ResNet & 73.61  \\ \hline
                    \end{tabular}

\end{table}

\item Using Different Number of Epochs: We have trained our model on different number of epochs as well. First we trained our model on 100 number of epochs and mean average precision was very low. Then we increased number of epochs up to 150. Now model was giving good than 100 number of epochs but it was still learning and validation loss was decreasing. So again we set number of epochs into 200 and it gives very good results. We also set number of epochs more than 200 and we set 250 and 300 number of epochs for training but our model precision was decreasing. So we set epoch value 200 which gives best result. 
\item Using Different Learning Rate Values: We also changed learning rate into different values to check at which value model learns best and gives best results. When we decreased learning rate value from 1e-5 to 1e-4 or even less then model was giving bad results. We also set learning rate value higher than 1e-5 and precision was not improving but it was decreasing. Learning rate value 1e-5 gives best results.  
\item Using Different IOU values: We have set different IOU value for non maximum suppression and check our results that at which value it is giving good results. Table 4.3 shows results at o.25, o.5 and o.75 IOU value. At o.25 IOU values model gives good results but it will not be accurate in real time scenario. Model will not give good results. Normally people set IOU value at 0.5 as standard to check that how model is accurately working.  
\end{enumerate}

\section{Conclusion \& Future Directions}
In this research work, we proposed a novel OAWD algorithm with application to detect weapons mainly guns and pistols along with orientation in visual data. OAWD is trained on our own created data set using horizontal bounding boxes and angle. The direction forecast is acted like a classification problem and regression problem by dividing every angle into eight different classes and by taking angle in the range from zero to one respectively. OAWD (classification) predicts bounding box offsets, object class, and angle class or every input object proposal. While OAWD (regression) predicts bounding box offset plus angle offset and object class. The proposed OAWD model not only provides orientation for some specific angle but also gives any angle of an object present in the range from zero degree to 180 degree. For training and assessment of the proposed model, a new weapon data set comprising of around 6400 annotated images have been gathered, which will soon be available publicly for further research work. The models we discussed in the literature review and compared model provides horizontal bounding boxes which have greater background information but our proposed model localizes better and provides oriented bounding boxes. OAWD is compared with Faster RCNN 
In a wide scope of tests, the proposed indicator has exhibited improved location and restriction execution for the undertaking of gun discovery.
        
In future work, the data set can be increased and more classes can be included. Different types of weapons like tanks, hand grenades, etc. can also be included in the data set. Every type of gun and riffle can be separately dealt with its own class. If we have 3-D data set for orientation then using deep learning techniques we can tell that either weapon is pointed towards humans or not. If we have such a data set that shows hidden objects then we can easily detect weapons and can assist law enforcement agencies for proper action in time. The accuracy of object detection can also be increased by rotating the feature map on a predicted angle and then again we use fully connected layers next to the rotated feature map for classifier and regression. Regression can also be introduced in Region proposal network module for rotated Region of interest at every angle.


%


%
%


\end{document}